\documentclass[nohyperref]{article}

\usepackage{microtype}
\usepackage{graphicx}
\usepackage{subfigure}
\usepackage{booktabs} 

\usepackage{hyperref}

\usepackage[accepted]{icml2022}

\usepackage{amsmath}
\usepackage{amssymb}
\usepackage{mathtools}
\usepackage{amsthm}

\usepackage[capitalize,noabbrev]{cleveref}

\theoremstyle{plain}
\newtheorem{theorem}{Theorem}[section]
\newtheorem*{hyp*}{Working Hypothesis}

\theoremstyle{definition}

\theoremstyle{remark}

\usepackage[textsize=tiny]{todonotes}

\icmltitlerunning{Enhancing Causal Estimation through Unlabeled Offline Data}

\begin{document}

\twocolumn[
\icmltitle{Enhancing Causal Estimation through Unlabeled Offline Data}



\icmlsetsymbol{equal}{*}

\begin{icmlauthorlist}
\icmlauthor{Ron Teichner}{yyy}
\icmlauthor{Ron Meir}{yyy}
\icmlauthor{Danny Eytan}{yyy,comp}
\end{icmlauthorlist}

\icmlaffiliation{yyy}{Department of Electrical Engineering, Technion - Israel Institute of Technology, Haifa, Israel}
\icmlaffiliation{comp}{Rambam Health Care Campus, Haifa, Israel}

\icmlcorrespondingauthor{Ron Teichner}{ron.teichner@campus.technion.ac.il}

\icmlkeywords{Machine Learning, ICML}

\vskip 0.3in
]




\printAffiliationsAndNotice{}  

\begin{abstract}

Consider a situation where a new patient arrives in the Intensive Care Unit (ICU) and is monitored by multiple sensors. We wish to assess relevant unmeasured physiological variables (e.g., cardiac contractility and output and vascular resistance) that have a strong effect on the patient’s diagnosis and treatment. We do not have any information about this specific patient, but, extensive offline information is available about previous patients, that may only be partially related to the present patient (a case of dataset shift). This information constitutes our prior knowledge, and is both partial and approximate. The basic question is how to best use this prior knowledge, combined with online patient data, to assist in diagnosing the current patient most effectively. Our proposed approach consists of three stages: \emph{(i)} Use the abundant offline data in order to create both a non-causal and a causal estimator for the relevant unmeasured physiological variables. \emph{(ii)} Based on the non-causal estimator constructed, and a set of measurements from a new group of patients, we construct a causal filter that provides higher accuracy in the prediction of the hidden physiological variables for this new set of patients. \emph{(iii)} For any new patient arriving in the ICU, we use the constructed filter in order to predict relevant internal variables. Overall, this strategy allows us to make use of the abundantly available offline data in order to enhance causal estimation for newly arriving patients. We demonstrate the effectiveness of this methodology on a (non-medical) real-world task, in situations where the offline data is only partially related to the new observations. We provide a mathematical analysis of the merits of the approach in a linear setting of Kalman filtering and smoothing, demonstrating its utility. 

%
\end{abstract}

\section{Introduction}
\label{sec:intro}
Physiological and biological systems, such as the cardiovascular system, are often approximated by simplified models. When real-time accurate estimation of system states is required, these models serve as a bridge between the obtained observations (e.g., heart-rate, blood-pressure, respiration rate) and the hidden system states (e.g., blood ventricle volumes). Due to the use of approximated models, the accuracy of estimation is lower than is potentially possible. A similar phenomenon occurs when a data-driven estimator is trained on a particular labeled dataset (i.e., one that includes the system's state), and then evaluated in a different environment, for example, a speech recognition machine trained in the United-States and evaluated in India is in-fact operating with respect to a misspecifed model. The problem of shifts between the training data and the deployment data is widely known in machine learning and referred to as \textit{dataset shift} \cite{quinonero2008dataset}.

Dataset shifts, and in particular dataset shifts in healthcare, pose a major concern since even slight deviations from the training conditions can result in wildly different performance \cite{subbaswamy2020development}. In recent years research on dataset shifts has focused on numerous directions among which are, identification and quantification of the limitations of a model when encountered with unseen data \cite{park2021reliable}, and, algorithms for learning models that guarantee stability against shifts \cite{subbaswamy2019preventing}. The majority of research in the field is concerned with models that process tabular data such as chest radiographs \cite{zech2018variable} or complete records of lab measurements \cite{caruana2015intelligible}. In the present study we consider dataset shifts on time-series data. The observed time-series evolves according to the dynamic rules of the observed system and, simultaneously, a filter estimates hidden unmeasured physiological variables. Our concern is in retaining the performance level of this real-time estimation in spite of the dataset shift. Our analysis takes advantage of dynamical properties of the data that are more robust to the shifts, and yields an algorithm that reduces the effect of a dataset shift.

To address the problem of causal estimation under a misspecifed model (a dataset shift), we introduce a novel model-based and data-driven estimation concept. Trained offline, the objective of a learned causal estimator is to reproduce the estimates obtained by a model-based non-causal estimator, while operating in a causal setup. This procedure, which fuses our prior knowledge with the new observed data, implicitly assumes that the non-causal estimates are reliable, and so poses a fundamental question - \textbf{is future information reliable in the context of estimation under a misspecified model setting?} 

To address this question, we first precisely formulate this problem and provide a methodological learning framework. Then, to analytically asses the reliability of the proposed methodology, in Section \ref{sec:prob} we derive novel performance bounds for a use case of a misspecified discrete-time Kalman model. In Section \ref{sec:example} we demonstrate the methodology on both a synthetic linear model and on real motion sensory data obtained from sensors embedded in smartphones. 
\section{Problem Setup}
We begin by describing a filtering problem (causal hidden state estimation) in general terms. The dynamics of a time-invariant system are described by
\begin{equation} \label{eq:dynamics}
    x_{k+1} = f(x_k, d_k, \omega_k) , 
\end{equation}
and an observation equation
\begin{equation} \label{eq:meas}
    z_k = g(x_k, d_k, v_k) , 
\end{equation}
where $x$ is the state of the system, $d$ is a known input, $\omega$ is a process noise term and $v$ is the measurement noise process. The index $k$ represents discrete time. 
A filter estimates the state $x_k$ based on past observations and inputs,
\begin{equation} \label{eq:est}
    \hat{x}_{k+1 \mid k} = F(z_{0:k}, d_{0:k}), 
\end{equation}
where $z_{0:k}$ ($d_{0:k}$) denotes all measurements (inputs) until, and including, time $k$. Formally this is a 1-step predictor, but, following \citet{anderson2012optimal}, we refer to it as a filter. It receives measurements according to a measurement equation \eqref{eq:meas} where $z$ is the received measurement. The objective of filtering \eqref{eq:est} is to produce a causal estimator that minimizes a cumulative cost function, $L_{F \mid Z} = \sum_{k=0}^{N-1}\operatorname{E}[{c_{k \mid k-1}}]$, where $c_{k \mid k-1}=C(x_k, \hat{x}_{k \mid k-1})$ is a cost term, for example, when minimizing mean-square-error, $C(\cdot)$ is given by $C(a,b)=\|a-b\|_2^2$.

In this study we consider the case where the measurements are generated by $\tilde{g}(\cdot)$
\begin{equation}\label{eq:inputReality}
    \tilde{z}_k = \tilde{g}(x_k,d_k,\tilde{v}_k),
\end{equation}
while the estimator erroneously assumes $g(\cdot)$ (and with $\tilde{v}_k$ a noise process independent of $v_k$). To illustrate, consider the task of geo-localization of a lecturer walking in an auditorium. In this case $x_k$ is the position and $z_k$ are acoustic measurements. The wrongly assumed measurement equation, $g(\cdot)$, is merely a simplified wave-propagation model that does not consider the highly complex (and non-available) wave-propagation pattern, $\tilde{g}(\cdot)$, of the indoor environment. As a result, the filter $F(\cdot)$ requires a rather long observation period before it produces an accurate estimation of $x_k$. Over the years, lecture recordings given in this auditorium were collected such that a dataset of recordings is available. We emphasize that the dataset is unlabeled in the sense that it does not include the system state, $x_k$, and ask, \textbf{under what conditions can an unlabeled offline dataset be exploited for obtaining an improved causal estimator?} Of particular interest to us is the case where there is a shift between past and present data, so that any previously learned model may be misspecified in the present setting.

Smoothing (non-causal state estimation) is an estimation method where the state estimate is based on both past and future observations and inputs, namely 
\begin{equation} \label{eq:estSmoother}
    \hat{x}_{k \mid N-1} = S(z_{0:N-1}, d_{0:N-1}), 
\end{equation}
and the objective is to minimize the cost function, $L_{S \mid Z} =\sum_{k=0}^{N-1}\operatorname{E}[{c_{k \mid N-1}}]$, where $c_{k \mid N-1}=C(x_k,\hat{x}_{k \mid N-1})$. In cases where the model is accurate, smoothing will have a lower estimation error compared to filtering \cite{simon2006optimal}, as is intuitively plausible. When the model is a simplified approximation of the system ($g(\cdot)$ instead of $\tilde{g}(\cdot)$) this is not always guaranteed. Yet, since the significant behaviours undoubtedly are modeled by the simplified model $g(\cdot)$, assuming that smoothing outperforms filtering on average is reasonable, and in many cases empirically true. The concept for learning an improved causal estimator is described in the following Working Hypothesis,
\begin{hyp*}\label{hyp:main}
    Consider the problem of estimating the hidden state of a system from observations, when an approximate system model is available , namely, we have an approximate state dynamics and observation model. Assume the following statements hold:
    \begin{enumerate} 
        \item A dataset of observations is available.
        \item Non-causal model-based smoothing outperforms causal filtering - in spite of model misspecification. 
        \item In addition to modeled behaviours, the observed data also displays unmodeled behaviours. These unmodeled behaviours are correlated, to some degree with the internal state of the system. 
    \end{enumerate}
    Then, a learned causal estimator trained on the dataset can obtain a lower error compared to model-based filtering.
\end{hyp*}
As an example of the Working Hypothesis consider a smoother trained on the simplified model of the cardiovascular system developed by \citet{zenker2007inverse}. This smoother operates on coarse time-averaged values of vital physiological signals (blood-pressure, heart rate, etc.) and, although it is able to distinguish between two cardiovascular shock types, it does so only by processing measurements collected over a long observation period that includes external perturbations to the system. On the other hand, the full raw wave pattern of the vital signals contains behaviours which are unmodeled by this simplified model. These unmodeled behaviours are correlated with the state of the cardiovascular system, and thus potentially allow learning a filter that is able to identify the type of shock earlier.
When the conditions of the Working Hypothesis hold we propose deriving an improved causal estimator by solving the optimization problem,
\begin{equation} \label{eq:estImprovedSol}
\begin{split}
    &\mathop{\mathrm{argmin}}_{\tilde{F}}\frac{1}{N} \operatorname{E}
        \left[ \sum_{k=0}^{N-1}  C\left(\tilde{x}_{k \mid k-1}, \hat{x}_{k \mid N-1}\right)\right]\\
        &\text{s.t.~} \tilde{x}_{k \mid k-1} = \tilde{F}(\tilde{z}_{0:k-1}, d_{0:k-1}), 
\end{split}
\end{equation}
where $\hat{x}_{k \mid N-1} = S(\tilde{z}_{0:N-1}, d_{0:N-1})$ is the non-causal estimator based on the misspecified mode, and $C(\cdot)$ is the application-specific cost function. Note that we measure the causal filter's estimator $\tilde{x}_{k \mid k-1}$ against the performance of the non-causal estimator $\hat{x}_{k \mid N-1}$.

\begin{figure}
    \centering
        \includegraphics[width=0.4\textwidth]{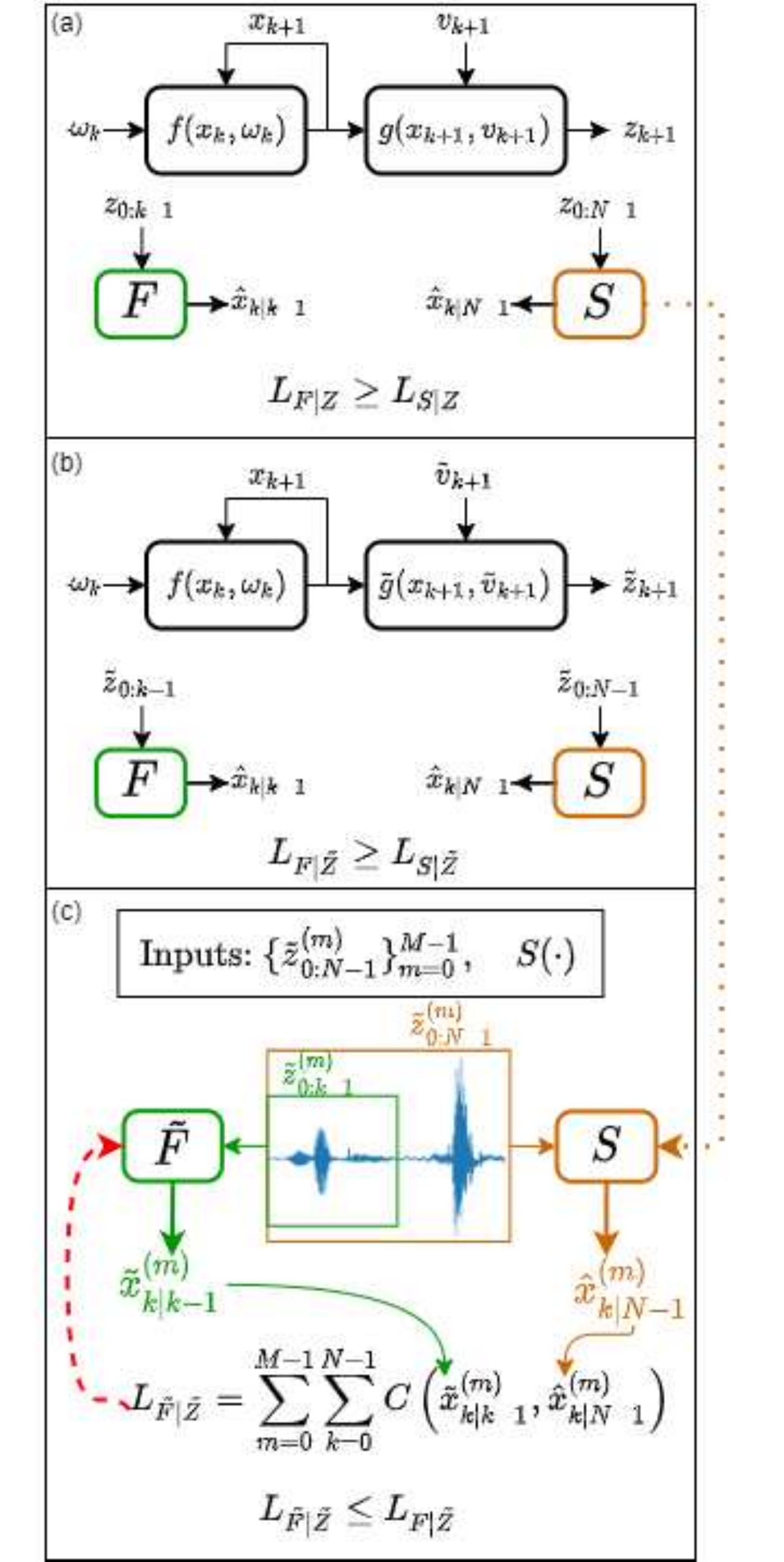}
        \caption{\label{fig:causalEstimator} \textbf{(a)} The system $f(\cdot)$ is driven by a process noise $\omega_k$ and has an internal state $x_k$. The observations $z_k$ are the output of the measurement equation $g(\cdot)$. The filter $F(\cdot)$ and smoother $S(\cdot)$ are matched to $(f,g)$ and therefore the filtering loss $L_{F \mid Z}=\sum_{k=0}^{N-1}\operatorname{E}\left[c_{k \mid k-1}\right]$ is higher than the smoothing loss, $L_{S \mid Z}=\sum_{k=0}^{N-1}\operatorname{E}\left[c_{k \mid N-1}\right]$. \textbf{(b)} In reality the observations are generated by $\tilde{g}(\cdot)$ while the estimators wrongly assume $g(\cdot)$. By the Working Hypothesis we assume that despite the misspecified measurement equation, the smoothing loss, $L_{S \mid \tilde{Z}}$ remains smaller than the filtering loss, $L_{F \mid \tilde{Z}}$. \textbf{(c)} Training setup of optimization \eqref{eq:estImprovedSol}. For all time-series in an offline dataset $\{\tilde{z}^{(m)}_{0:N-1}\}_{m=0}^{M-1}$, the model-based smoother, $S(\cdot)$, estimates the state $x_k$ using the complete observed time-series, $\tilde{z}_{0:N-1}$. The learned filter, $\tilde{F}(\cdot)$, is trained to reproduce the smoother's estimations while only observing $\tilde{z}_{0:k-1}$ and obtains a loss $L_{\tilde{F} \mid \tilde{Z}}$ smaller than the loss of the original filter, $L_{F \mid \tilde{Z}}$.}
\end{figure}

\textbf{Methodology} Our methodology is summarized in figure \ref{fig:causalEstimator} where (a) depicts the filter $F(\cdot)$ and smoother $S(\cdot)$ which are matched to a model given by $f(\cdot)$ and $g(\cdot)$ (the input term $d_k$ was omitted for clarity). The smoothing loss, $L_{S \mid Z}=\sum_{k=0}^{N-1}\operatorname{E}\left[ c_{k \mid N-1}\right]$, is lower than the filtering loss $L_{F \mid Z}=\sum_{k=0}^{N-1}\operatorname{E}\left[  c_{k \mid k-1}\right]$. In (b) we depict the operation of these estimators in case where the measurements are in fact generated by $\tilde{g}(\cdot)$ due to a misspicification in the measurement equation. By the Working Hypothesis we assume that despite this, the smoothing loss, $L_{S \mid \tilde{Z}}$, remains smaller than the filtering loss, $L_{F \mid \tilde{Z}}$. In (c) we depict the training setup of optimization \eqref{eq:estImprovedSol}. For all $M$ time-series in an offline dataset $\{\tilde{z}^{(m)}_{0:N-1}\}_{m=0}^{M-1}$, the smoother $S(\cdot)$ estimates the state $x_k$ using the complete observed time-series, $\tilde{z}_{0:N-1}$. The learned filter, $\tilde{F}(\cdot)$, is trained to reproduce the smoother's estimations while only observing $\tilde{z}_{0:k-1}$. 

This methodology is suited to numerous different use-cases where simplified models are assumed and abundant off-line data is available. For example, earthquake predictions from seismic measurements \cite{ogiso2021estimation}, self localization by GPS in an urban environment \cite{merry2019smartphone} and the above-mentioned simplified physiological models that fail to accurately reproduce the fine details of physiological signals \cite{Keener:1998:MP:289685}. 
\subsection{Related work}
Deep learning tools are widely used in inference problems, both causal and non-causal. A large line of work concerns the inference of a latent state space model (e.g.,  \citet{karl2016deep,krishnan2017structured}). While latent variables in such methods help in understanding the underlying dynamics of a system, they are usually not variables of predefined physical or physiological meaning such as the variables we would like to estimate. A second line of work concerns in the inference of predefined state variables using deep learning tools (see \citet{ambrogioni2017estimating, ni2021rtsnet}), yet, these frameworks require a labeled dataset containing the states of the system while our method requires the measurements only.

Machine Learning for healthcare is extremely vulnerable to dataset shifts which widely occur due to demographic diversity and different treatment protocols. To build models which are robust to these shifts \citet{subbaswamy2019preventing} analyse the data generating process, identify expected changes and perform sophisticated data augmentation in training. In \citet{subbaswamy2021evaluating} model robustness to dataset shifts is analyzed to validate its use in a different deployment environment. While most studies, both on training protocols and on model robustness analysis, concern tabular data, \citet{maldonado2021out} consider time-series data and suggest a robust training method. Our approach differs in that our methodology handles an \emph{existing} reduction in performance due to a dataset shift, and suggests an algorithm that benefits from the relative robustness of non-causal estimation to dataset shifts.

In Section \ref{sec:prob} we derive novel performance bounds for the case of a misspecified discrete Kalman model by considering an adversarial input. This resembles the development and analysis of the $H_\infty$ filter and smoother \cite{zames1981feedback, yaesh1991transfer, hassibi1999indefinite,mirkin2003h,simon2006optimal, 4287144}, where an adversarial behavior of the process and measurement noises is considered such that the objective of the adversary is to maximize either the filtering or the smoothing error. We deal with a different objective, namely, determining the largest degree of misspecification for which smoothing still outperforms filtering, in accordance with the Working Hypothesis. In addition, $H_\infty$ does not constrain the energy of the adversary while we do.
%

\section{Performance bounds in a linear setup}\label{sec:prob}
\subsection{Linear problem formulation}
As a special use case, we analytically investigate the case where the dynamics and the measurement equations are linear, and the estimators are the celebrated Kalman filter and smoother e.g., \cite{anderson2012optimal}. To obtain smoothing vs. filtering performance bounds we consider the difference between $g(\cdot)$ and $\tilde{g}(\cdot)$ to be caused by an adversarial player (termed \emph{adversary}).
Assuming a constrained adversary energy, we developed bounds on the decrement in the performance of smoothing with respect to filtering. The bounds are easily obtained by solving scalar convex optimization problems.
Assume the discrete-time, linear, finite-dimensional, time-invariant, asymptotically stable system, 
\begin{equation}\label{eq:sys}
    \begin{split}
        {x}_{k+1} &= A{x}_k + \omega_k, \quad
        {z}_{k} = H'{x}_k + v_k,
    \end{split}
\end{equation}
with $x_k \in {\mathbb{R}}^n$, $z_k \in {\mathbb{R}}^m$. The noise terms $v_k$ and $\omega_k$ are independent, zero mean, stationary Gaussian white processes, $\operatorname{E}[v_k v_l'] = R \delta_{kl},
    \operatorname{E}[\omega_k \omega_l'] = Q \delta_{kl}$. 
We assume that $A^{-1}$ exists which is the case if $A$ arises from a real system because then $A$ is the result of a matrix exponential that is always invertible \cite{simon2006optimal}. The objective of filtering, as described in Section \ref{sec:intro}, is to produce a causal estimator that minimizes a cost function, $L_{F \mid Z} = \sum_{k=0}^{N-1}\operatorname{E}[{c_{k \mid k-1}}]$, where here $c_{k \mid k-1} = \|e_{k \mid k-1}\|_2^2$ and $e_{k \mid k-1}$ is the estimation error defined by $e_{k \mid k-1} \overset{\triangle}{=} {x}_{k} - \hat{x}_{k \mid k-1}$. Suppose that for the estimation of the state ${x}_{k+1}$ (for times $k \geq 0$) based on the measurements ${z}_{0:k}$, the time invariant Kalman filter is used,
\begin{equation} \label{eq:filter}
    \begin{split}
        \hat{x}_{k+1 \mid k} &= \tilde{A}\hat{x}_{k \mid k-1} + K{z}_k, \quad
        \hat{x}_{0 \mid -1} = \hat{x}_0,
    \end{split}
\end{equation}
with $\tilde{A} = A-KH'$, $K=A \bar{\Sigma}H(H' \bar{\Sigma} H + R)^{-1}$ and $\bar{\Sigma}$ is the solution of the steady-state Riccati equation,
\begin{equation} \label{eq:barSigma}
    \bar{\Sigma} = A[\bar{\Sigma} - \bar{\Sigma}H(H' \bar{\Sigma} H + R)^{-1}H'\bar{\Sigma}]A' + Q.
\end{equation}
We note that for the time invariant asymptotically stable system \eqref{eq:sys} there exists a solution $\bar{\Sigma}$ to equation \eqref{eq:barSigma} that yields $|{\lambda_i(\tilde{F})}| < 1$ (for all $i$), guaranteeing that the filter \eqref{eq:filter} is asymptotically stable. For a detailed analysis of time-invariant Kalman filters see Chapter 4 in \cite{anderson2012optimal}.

The objective of smoothing is to produce a non causal estimator that minimizes a cumulative cost function, $L_{S \mid Z} =\sum_{k=0}^{N-1}\operatorname{E}[{c_{k \mid N-1}}]$, where $c_{k \mid N-1} = \|e_{k \mid N-1}\|_2^2$ and $e_{k \mid N-1}$ is the estimation error defined $e_{k \mid N-1} \overset{\triangle}{=} {x}_{k} - \hat{x}_{k \mid N-1}$. Suppose that for a non-causal estimation of the state $x_k$ based on the measurements $z_{0:N-1}$ the Kalman smoother is used \cite{anderson2012optimal}, 
\begin{equation} \label{eq:smoother}
    \begin{split}
        \hat{x}_{k \mid N-1} &= \left(\bar{\Sigma} {\tilde{A}}' {\bar{\Sigma}}^{-1}\right) \hat{x}_{k+1 \mid N-1}\\
        &+ \left(A^{-1} - \bar{\Sigma} {\tilde{A}}' {\bar{\Sigma}}^{-1}\right) \hat{x}_{k+1 \mid k},
    \end{split}
\end{equation}
where the initial condition is,
\begin{equation} \label{eq:smootherInit}
    \begin{split}
        &\hat{x}_{N-1 \mid N-1} = \Bar{\Sigma} [A \Bar{\Sigma}]^{-1}\tilde{x}_{N \mid N-1}\\
        &- (\Bar{\Sigma} [A \Bar{\Sigma}]^{-1}\tilde{A} + I - \Bar{\Sigma} [A \Bar{\Sigma}]^{-1} KH')\hat{x}_{N-1 \mid N-2}.
    \end{split}
\end{equation}
We consider a case in which starting at $k=0$, an additive input $u_k \in {\mathbb{R}^n}$ (termed adversary, unknown to the estimators) enters the filter such that $\tilde{z}_k = {z}_k + H'u_k$,
so that the filter receives $\tilde{z}_k$ instead of ${z}_k$, and produces output
\begin{equation} \label{eq:filterAdvInput}
    \hat{x}_{k+1 \mid k} = \tilde{A}\hat{x}_{k \mid k-1} + K\tilde{z}_k.
\end{equation}
We analyze the difference in estimation error of smoothing vs filtering due to their erroneous observation model, and establish a bound on the difference in mean error energies, assuming limited adversary energy.
\begin{equation*}
    L_{F \mid Z} - L_{S \mid Z} = \operatorname{E} \left[ \sum_{k=0}^{N-1} \|{e_{k \mid k-1}}\|_2^2 - \|{e_{k \mid N-1}}\|_2^2 \right]~.
\end{equation*}
Then, the optimization problem is
\begin{align}
    &I_N \overset{\triangle}{=}\min_{u_{0:N-1}} \frac{1}{N} \left(L_{F \mid Z} - L_{S \mid Z}\right)\nonumber\\
        %
        %
        &\text{s.t. } \frac{1}{N} \sum_{k=0}^{N-1} \|{u_k}\|_2^2 \leq \gamma, \quad \gamma \geq 0.
        \label{eq:optObj}
\end{align}
Note that the expectation is with respect to the \emph{true} model statistics, not the one assumed – incorrectly – by the estimators. Further, note that \eqref{eq:optObj} is a non-convex optimization problem of dimension $nN$ (while minimizing a sum of norms subject to a quadratic convex constraint is a convex optimization problem, we are looking for minimizing the difference between two norm sums, leading to a non-convex problem \cite{boyd2004convex}). To obtain a worst-case bound, we assume the adversary has full knowledge about the system and the estimator as well as access to all past and future states and measurement noises. In cases where $I_N > 0$, and following the Working Hypothesis, an offline unlabeled dataset can be exploited for learning an improved causal estimator, as defined in optimization problem \eqref{eq:estImprovedSol} and depicted in figure \ref{fig:causalEstimator}.

\textbf{Notation} Our derivations are compactly expressed with the help of block matrices and vectors. We generally define a block-matrix $A$ of dimensions $(N \times n) \times (M \times n)$ such that $A[r,c]$, with $0 \leq r < N$ and $0 \leq c < M$ is the $n \times n$ matrix that lies between rows $nr$ to $n(r+1)-1$ and columns $mc$ to $m(c+1)-1$ of the block-matrix $A$ and by $A[r,c][i,j]$ the entry in row $i$ column $j$ of the $n \times n$ matrix $A[r,c]$.\\
For a symmetric and full rank matrix $A$ we denote the real eigenvalues by $\lambda_i(A)$ and by $v_i(A)$ the corresponding unit-length eigenvector. We define $\lambda_{\operatorname{max}}(A)$ to be the largest eigenvalue of $A$. We compactly denote all vectors $\{x_i, x_{i+1}, \dots, x_j\}$ with $j \geq i$ as $x_{i:j}$.
We define $A'$ to be the transpose of the matrix $A$ and ${\mathcal I}(\cdot)$ to be the indicator function.

Let $\tilde{e}_{k \mid k-1}$ and $\tilde{e}_{k \mid N-1}$ be the errors of the Kalman filter and smoother respectively in the absence of an adversary, i.e. $u_k \equiv 0$. We refer them as the \emph{nominal errors}. For simplicity and without loss of generality we assume $\tilde{e}_{0 \mid -1} \sim \mathcal{N}(0, \bar{\Sigma})$. From the linearity of the estimators we deduce for all\\$0 \leq k < N$,
\begin{equation}\label{eq:linearErr}
    \begin{split}
        e_{k \mid k-1} = \tilde{e}_{k \mid k-1} - \xi_{k}, \quad
        e_{k \mid N-1} = \tilde{e}_{k \mid N-1} - \xi^{s}_{k},
    \end{split}
\end{equation}
with $\xi_{k}$ and $\xi^{s}_{k}$ being the outputs of the filter and the smoother respectively if the adversary was the only input, i.e. $z_k \equiv 0$ (see Appendix \ref{sec:kalblock}),
\begin{equation}\label{eq:kalmanOutput}
    \begin{split}
        \xi_{k} &= \sum_{i=0}^{k-1}\tilde{A}^{k-i-1}KH'u_{i} \quad \forall{k > 0}\\
        \xi^s_{k} &=  \sum_{i=0}^{k-1} \tilde{B}_{k,i} KH'u_i
        + \sum_{i=k}^{N-1} \tilde{C}_{k,i} KH'u_i,
    \end{split}
\end{equation}
and
\begin{equation*}
    \begin{split}
        &\tilde{B}_{k,i} =  \tilde{A}^{k-i-1}
        - \Bar{\Sigma}  \tilde{D}_{k,i,k} \\
        &\tilde{C}_{k,i} = \Bar{\Sigma} (\tilde{A}^{i-k'} [A \Bar{\Sigma}]^{-1} 
        - \tilde{D}_{k,i,i+1})\\
        &\tilde{D}_{k,i,m} = \sum_{n=m}^{N-1} \tilde{E}_{k,i,n}\\
        &\tilde{E}_{k,i,n} = \tilde{A}^{n-k'} [A \Bar{\Sigma}]^{-1} K H'\tilde{A}^{n-i-1}.
    \end{split}
\end{equation*}
Let $\tilde{e}_N, \tilde{e}^s_N, \tilde{\xi}_N, \tilde{\xi}^s_N, u_N \in {\mathbb{R}^{nN}}$ be the block vectors
$\tilde{\xi}_N = \begin{bmatrix}
    \xi_0' \dots \xi_{N-1}',
\end{bmatrix} '$, 
$\tilde{\xi}^s_N = \begin{bmatrix}
    \xi^{s'}_0 \dots \xi^{s'}_{N-1},
\end{bmatrix} '$, 
$ \tilde{e}_N = \begin{bmatrix}
    \tilde{e}_{0 \mid -1}' \dots \tilde{e}_{N-1 \mid N-2}'
\end{bmatrix} '$, 
$\tilde{e}^s_N = \begin{bmatrix}
    \tilde{e}_{0 \mid N-1}' \dots \tilde{e}_{N-1 \mid N-1}'
\end{bmatrix} '$, 
$u_N = \begin{bmatrix}
    u_0' \dots u_{N-1}',
\end{bmatrix} '$
and let $\bar{\Xi}_{N}$ and $\bar{\Xi}^s_{N}$ be the block matrices of dimensions $[nN \times nN]$ having the entries,
\begin{equation*}
    \begin{split}
       \bar{\Xi}_{N}[r,c] = {\mathcal I}(c < r) \tilde{A}^{r-1-c}KH', \quad
        \bar{\Xi}^s_{N}[r,c] = {\tilde{G}_{r,c}}KH',
    \end{split}
\end{equation*}
with $\tilde{G}_{k,i} = {\cal I}(i<k){\tilde{B}_{k,i}} + {\cal I}(i \geq k){\tilde{C}_{k,i}}$.
The matrices $\bar{\Xi}_{N},\bar{\Xi}^s_{N}$ are known as the \emph{transfer operator} of the time-invariant Kalman filter and the corresponding Kalman smoother which relates the outputs to the inputs by $\tilde{\xi}_N = \bar{\Xi}_{N} u_N$ and $\tilde{\xi}^s_N = \bar{\Xi}^s_{N} u_N$ (see Appendix \ref{sec:kalblock} for a detailed derivation of the block operators). Using the block notation we rewrite optimization problem \eqref{eq:optObj} as
\begin{align}\label{eq:optObjBlock}
        &I_N(\gamma) = 
        \Delta\alpha_N
        + \operatorname{min}_{\|{u_N}\|_2^2 \leq 1} \operatorname{E}[{u_N}' B_N {u_N} + 2{b_N}' {u_N}]
        %
\end{align}
with $\Delta\alpha_N=\alpha_N - \alpha^s_N$, 
$B_N = \gamma(\Xi_N - \Xi^s_N)$, 
$b_N = -\sqrt{\frac{\gamma}{N}}({\bar{\Xi}_N}'{\tilde{e}_N} - {\bar{\Xi}^{s'}_N}{\tilde{e}^s_N} )$, %
$\Xi_N={\bar{\Xi}_{N}}'\bar{\Xi}_{N}$, 
$\Xi^s_N={\bar{\Xi}^{s'}_{N}}\bar{\Xi}^s_{N}$ %
and where $\alpha_N\overset{\triangle}{=}\frac{1}{N} \operatorname{E}[{\|{\tilde{e}_N}\|_2^2}]$ and $\alpha^s_N\overset{\triangle}{=}\frac{1}{N} \operatorname{E}[{\|{\tilde{e}^s_N}\|_2^2}]$ are the mean error energy in the absence of an adversary. Denote by ${u^{*}_N}$ the optimal solution for the adversary, then, the bound is always of the form
\begin{equation}\label{eq:Bn}
    \begin{split}
    &I_N(\gamma) = 
        \Delta\alpha_N
        + \operatorname{E}[ {u^{*}_N}' B_N {u^{*}_N} + 2 {b_N}' {u^{*}_N}].
    \end{split}
\end{equation}
We note that the bound on the difference between smoothing and filtering mean error energies is the sum of the difference in nominal error energies, $\Delta\alpha_N$, and a quadratic term that is a function of the adversarial strategy $u_N$ and the nominal errors. For all $N$ and $\gamma \geq 0$ it clearly holds that $\Delta\alpha_N \geq I_N(\gamma)$.
%
%
%
%
%
\subsection{Linear problem performance bound}
%
%
As seen in \eqref{eq:optObjBlock} the optimal adversary strategy, which yields the lower bound on the improvement of non-causal estimation is a function of the nominal errors $\tilde{e}_N$ and $\tilde{e}^s_N$. In the following theorem, we formulate the smoothing vs filtering performance in a specific finite-horizon instance. The mean improvement of smoothing as a function of the unmodeled behavior energy $\gamma$ is obtained by averaging multiple independent finite-horizon instances.
\begin{theorem}\label{thm:th3}
Let ${\lambda}^*$ be the solution to the scalar convex optimization problem,
\begin{equation}\label{eq:scalarConvex}
    \begin{split}
        &{\lambda}^* = \operatorname{argmin}_\lambda \sum_{i=0}^{n-1} \frac{({v_i'({B_N})} b_N)^2}{\lambda_i({B_N}) + \lambda} + \lambda\\
            &\mathrm{s.t.} \lambda \geq \operatorname{max}(-\lambda_{\operatorname{min}}(B_N),0).
    \end{split}
\end{equation}
Then, in a specific finite horizon case of length $N$ the optimal strategy is $u^*_N=-\sqrt{N}(B_N + \lambda^* I)^{-1}b_N$ and the bound $I_N(\gamma)$ is obtained by evaluating equation \eqref{eq:Bn}.
\end{theorem}
When the unmodeled behavior energy is on average (per time-step) $\gamma$ and $I_N(\gamma) > 0$, an improved causal estimator, in the sense of a lower mean-square-error, can be learned. Section \ref{sec:example} provides an example in which utilizing deep-learning tools an improved estimator is obtained. 

At first sight, solving problem \eqref{eq:optObjBlock} seems unachievable due to its non-convexity, yet, this optimization problem can be trivially converted to the form of the following problem, \eqref{eq:boydPrimal}, which satisfies strong duality with the scalar convex optimization problem, \eqref{eq:scalarConvex}. Consider the problem of minimizing a quadratic function over the unit ball,
\begin{equation}\label{eq:boydPrimal}
    \begin{split}
        &\operatorname{minimize}_{\|u_N\|_2^2 \leq 1} u_N' B_N u_N + 2b_N' u_N
        %
    \end{split}
\end{equation}
where $B_N$ is a symmetric matrix and $b_N \in {\mathbb R}^n$ \cite{boyd2004convex}. We do not assume that $B_N$ is positive-semi-definite (PSD) so problem \eqref{eq:boydPrimal} is potentially a non-convex optimization problem. The dual problem is optimization problem \eqref{eq:scalarConvex}. Strong duality holds for problem \eqref{eq:boydPrimal} and thus its optimal value can be obtained by solving the scalar convex optimization problem \eqref{eq:scalarConvex}.

We refer the reader to Chapter 5 in \citet{boyd2004convex} for a detailed overview on Duality in optimization, as well as to a proof (equation 5.28 and Appendix B in \citet{boyd2004convex}) of the existence of the property of strong duality between optimization problems \eqref{eq:boydPrimal} and \eqref{eq:scalarConvex}.
\section{Examples}\label{sec:example}
In this section we first utilize the suggested methodology on a synthetic linear example for which we can calculate the bound presented in Theorem \ref{thm:th3}. Then, in Section \ref{sec:HARexample}, we utilize the methodology to enhance real-time estimations of human activity from motion sensory data obtained from sensors embedded in smartphones \cite{stisen2015smart}. 
Code for both examples is available at\href{https://github.com/RonTeichner/Enhancing-causal-estimations}{GitHub}
\subsection{Synthetic example}
\begin{figure}
    \centering 
        \includegraphics[width=0.4\textwidth]{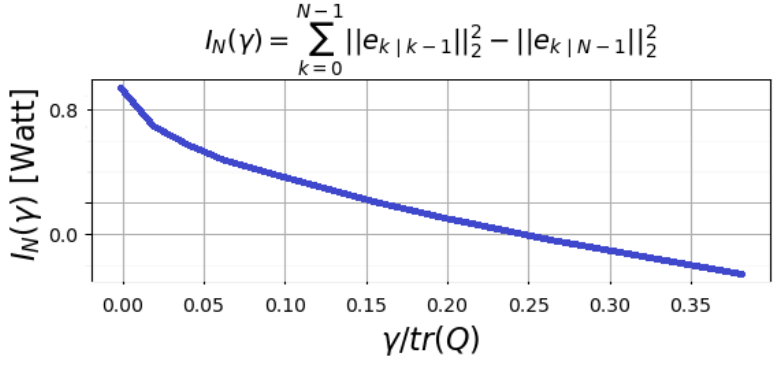}
        \caption{\label{fig:gammaPlot} Evaluating the bound of Theorem \ref{thm:th3} for system \eqref{eq:2dim_systemExampleFilteringBounds} with $N=50$. Smoothing is guaranteed to outperform filtering, $I_N(\gamma)>0$, for unmodeled behavior energy of up to $25\%$ of the (per time-step) input energy to the system, $\mathrm{tr}(Q)$.  }
\end{figure}
\begin{table}\small
\caption{Mean square error [Watts] for accurate $(u_k \equiv 0)$ and approximated models for system \eqref{eq:2dim_systemExampleFilteringBounds}.}
\begin{tabular}{||c c c||} 
 \hline
 Estimator & accurate & approximated  \\ [0.5ex] 
 \hline\hline
 Kalman filter & 1.86 & 2.54 (+36\%)  \\ 
 Kalman smoother & 0.92 & 1.17 (+27\%) \\
 Learned filter & 1.86 & 2.14 (+15\%) \\ [1ex]
 \hline
\end{tabular}
\label{table:sim}
\end{table}
Given a linear dynamical system we would like to calculate, by using Theorem \ref{thm:th3}, the maximal value of the mean unmodeled energy $\gamma$ (per time-step) for which it is still guaranteed that the smoothing error is lower than the filtering error, $I_N>0$, and therefore, by the Working Hypothesis there exists a potential for learning an improved causal estimator by the training procedure depicted in figure \ref{fig:causalEstimator}. Consider the system 
\begin{equation}\label{eq:2dim_systemExampleFilteringBounds}\small
    B = \begin{bmatrix}
        -0.898 & 0.950 \\ -0.056 & 0.569
    \end{bmatrix}, 
    \quad 
    H = \begin{bmatrix}
        0.443 & 0.862 \\ -0.220 & -0.100
    \end{bmatrix},
\end{equation}
where each matrix element was drawn randomly from a standard normal distribution,  and the matrices were scaled such $\lambda_{\max}(A)<1$ and $\|H\|_2=1$), with noise covariance matrices $Q = R = 0.5I$. Figure \ref{fig:gammaPlot} depicts the smallest guaranteed improvement of Kalman smoothing over Kalman filtering for different values of $\gamma$. For unmodeled behavior energies of $\gamma < 0.25 \mathrm{tr}(Q)$, non-causal estimation is guaranteed to outperform causal estimation and so there is a potential of learning an improved causal estimator.

Since $I_N(\gamma)$ is a worst case bound, in practice smoothing outperforms filtering for higher unmodeled energies than indicated in figure \ref{fig:gammaPlot}. To emphasize this, consider the non-linear unmodeled behavior,
\begin{equation*}
u_{k,0} = {\mathcal I}(x_{k,1} > 0 )x_{k,1}^2, \quad 
u_{k,1} = {\mathcal I}(x_{k,0} > 0 )x_{k,0}^2.
\end{equation*}
where $x_{k,i}$($u_{k,i}$) is the $i$\textsuperscript{th} entry of $x_k$($u_k$) and in which $\gamma$ is measured to be $12.4 \mathrm{tr}(Q)$. The estimators receive $\tilde{z}_k=z_k+H'u_k$ instead of $z_k$ 
which they are optimal for. We learned a causal estimator using the training setup depicted in figure \ref{fig:causalEstimator}. The learned filter is a recurrent neural network made up of two LSTM layers each with a hidden size $h=10$ \cite{zhang2020dive}. The input is $z_k \in {\mathbb{R}}^2$ and the output of the RNN, $h_k \in {\mathbb{R}}^h$ is connected to a linear layer whose output is $\tilde{x}_{k+1 \mid k} \in {\mathbb{R}}^2$. The RNN is trained on batches of size $b=320$ using the Adam optimizer \cite{kingma2014adam}, with learning rate $10^{-3}$. 

Table \ref{table:sim} lists the mean square errors, where, as a sanity check, we also learned an estimator for the case $u_k \equiv 0$. Due to the presence of the unmodeled behavior, the mean-square-error of the learned estimator increased by $15\%$ whereas the Kalman filter suffered an increase of $36\%$, indicating that the learned causal estimator successfully extracted features of the unmodeled behavior $u_k$, even when the model was misspecified. The values in table \ref{table:sim} were averaged over 10000 evaluations of the learned estimator on independent time-series leading to an accuracy of $\pm 0.13\%$. To assess the stability of the training procedure we repeated the procedure process $100$ times on independent datasets. The performance of an estimator learned via our training procedure has a standard deviation value of $\pm 0.28\%$. We note that similar results were obtained for other model systems with similar characteristics. 
\subsection{Enhancing real-time human activity recognition}\label{sec:HARexample}
We evaluated our proposed methodology to enhance real time estimation of human activity from motion sensory data. Human Activity Recognition (HAR) is the estimation of motion activity (standing, biking, ect.) from motion sensory data. HAR is based on the assumption that specific body movements translate into characteristic sensor signal patterns which can be sensed and classified. Off-the-shelf smartphones readily support numerous embedded sensors such as accelerometer, gyroscope and compass with the accelerometer being one of the earliest and most ubiquitous. Accelerometer measurements allow for the recognition of a wide variety of human activities \cite{casale2011human, bayat2014study, ignatov2018real}, capabilities which are integrated into different mobile applications such as the \href{https://www.sony.com/ke/electronics/truly-wireless/wf-1000xm3}{SONY WF-1000XM3} earbuds mobile application.

Different smartphones use different inbuilt accelerometer models which differ in precision, gains, resolution, biases, sampling rate heterogeneity and sampling rate instabilities \cite{stisen2015smart}. \citet{dey2014accelprint} state that smartphones are often well distinguishable by their accelerometer fingerprint and \citet{8626081} even suggest a fingerprint based authentication scheme.

In 2015, a systematic study of heterogeneity in motion-based sensing and its impact on HAR was performed by \citet{stisen2015smart} who gathered sensory data. The dataset, \textit{``Heterogeneity Human Activity Recognition Dataset"}, is publicly \href{http://archive.ics.uci.edu/ml/datasets/heterogeneity+activity+recognition}{available}. In this study, a total of 9 users carried different smartphone models while following a scripted set of activities. External differentiating factors were minimized by keeping all devices in a tight pouch carried by the users around their waist and by keeping the smartphone's CPU usage at a minimum.

\begin{figure}
    \centering
        \includegraphics[width=0.5\textwidth]{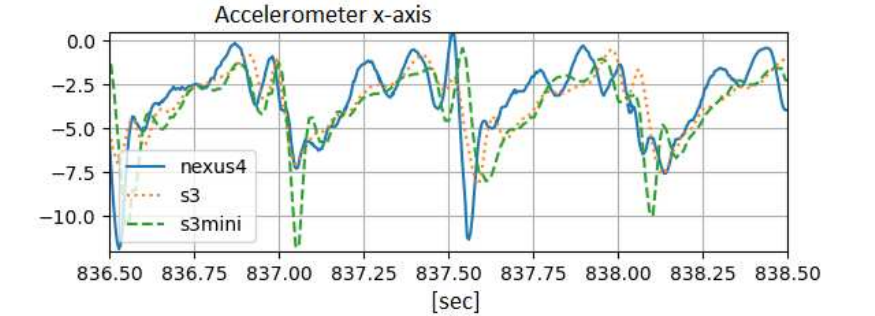}
        \caption{\label{fig:Acc_x} Accelerometer measurements retrieved simultaneously from Nexus4 (blue), S3 (orange) and S3mini (green) smartphones while user's activity is walking \cite{stisen2015smart}}
\end{figure}

In this setup the same dynamics is measured by each sensor and so the differences between obtained measurements originate only from the dissimilarities between the sensors, thus, by definition, a different measurement equation is the source of dissimilarities between measurements obtained from different devices. Figure \ref{fig:Acc_x} depicts accelerometer measurements retrieved from three different devices carried simultaneously by the same user.

For a specific device we define the data generating model as $M=(f,g)$ and the matched filter and smoother by $F$ and $S$, respectively. As is intuitively plausible, the loss of the smoother is guaranteed to be smaller than the loss of the filter, $L_{F \mid Z} \geq L_{S \mid Z}$, as illustrated in figure \ref{fig:causalEstimator}-a.

A second device (consider for example the first is a \href{https://www.samsung.com/ie/smartphones/others/galaxy-s3-mini-i8190-gt-i8190rwn3ie/}{Samsuing-S3mini} and the second an \href{https://www.lg.com/ae/smartphones/lg-Nexus-4-E960}{LG-Nexus4}) outputs measurements $\tilde{z}_k$ according to a model $\tilde{M}=(f,\tilde{g})$. The filter $F$ is sub-optimal when operating on measurements $\tilde{z}_k$ due to the differences between the models $M$ and $\tilde{M}$ which introduce a dataset shift.%
We tackle the scenario where we would like to derive a dedicated filter $\tilde{F}$ but the model $\tilde{M}$ is unknown and labeled data (measurements coupled with the corresponding states) is unavailable. Instead, abundant offline data is available, containing only the measurements $\tilde{z}_{k}$. By the Working Hypothesis we argue that despite the model mismatch, smoothing outperforms filtering, $L_{F \mid \tilde{Z}} \geq L_{S \mid \tilde{Z}}$, see figure \ref{fig:causalEstimator}-b. This superiority of smoothing allows us to learn an improved filter $\tilde{F}$ for the second device via optimization \eqref{eq:estImprovedSol} as illustrated in figure \ref{fig:causalEstimator}-c.

The observations ($z_{k},\tilde{z}_k \in {\mathbb{R}}^3$) are three dimensional accelerometer measurements and, while the state $x_k$ contains the full state-space description of motion dynamics, we are interested only in estimating the entry which represents the type of activity which, in this dataset takes one of six possible values. We thus view the state as $x_k \in \{a_i\}_{i=0}^5$ where the components $\{a_i\}$ correspond to sitting, standing, walking, stairs-up, stairs-down and biking. Therefore, we design the output $\hat{x}_{k \mid k-1}$ ($\hat{x}_{k \mid N-1}$) of a filter (smoother) to be a distribution over the categories $a_i$.

\begin{figure}
    \centering 
        \includegraphics[width=0.5\textwidth]{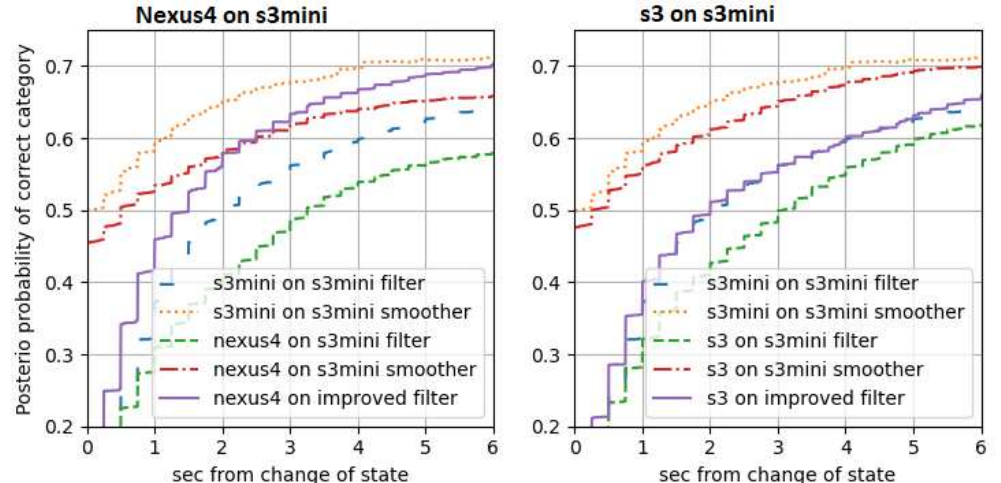}
        \caption{\label{fig:Improving}Left - Posterior probability assigned to the correct category after a change in activity state. The posterior of s3mini by matching filter $F$ (blue) and smoother $S$ (orange); Degraded performance when $F$ operates on LG-Nexus4 (green); In accordance with the Working Hypothesis, better performance is obtained by the learned filter $\tilde{F}$ (purple) that was trained on estimations of $S$ (red). Right - same procedure for s3 evaluated on s3mini.}
\end{figure}
For all estimators we utilize the DeepConvLSTM architecture presented in \cite{ordonez2016deep} with slight changes. See Appendix \ref{sec:HarTech} for complete technical details. We evaluate the performance of a filter by the posterior probability assigned by it to the correct ground-truth category, $P(\hat{x}_{k \mid k-1} = x_k \mid z_{0:k-1})$. As we are interested in real-time estimation, we examine the posterior probability in the first few seconds after a change in the activity state has occurred. Table \ref{table:simHAR} displays the performance of filters and smoothers trained on one smartphone and evaluated on a \emph{different} smartphone. In all cases, smoothing outperforms filtering as suggested by the Working Hypothesis. For the two worst performing unmatched filters (in bold text in table \ref{table:simHAR}) we learned improved filters by training them to reproduce the estimations of the (unmatched) smoothers. Figure \ref{fig:Improving} depicts the average performance of $F$ on measurements generated by $M$ or $\tilde{M}$ as well as the performance of $\tilde{F}$ on $\tilde{M}$. The results were averaged over $10$ independent evaluations of the methodology on different train/test splits.

The reported results display a successful evaluation of the proposed methodology, where causal estimation was enhanced by utilizing an unlabeled offline dataset. We note that our work lacks a common ground for comparison with existing works on HAR. Training estimators on just a single phone model, or on the outputs of a non-causal estimator is a unique setup. While state-of-the-art performance reported on HAR was obtained by training on all available smartphone-data together with sophisticated data augmentation techniques, we did not use any data augmentation so as to test the bare concept of causal estimation enhancement.

\begin{table} \small
\caption{Posterior probability assigned to the correct category two seconds after a change in activity state has occurred. First column lists the tested phone and second column the phone the estimators are matched to. Smoothing outperforms filtering in all cases. Figure \ref{fig:Improving} depicts the complete posterior probability vs time curves for the two worst performing filters (in bold).}
\begin{tabular}{||c c c c||} 
 \hline
 Phone & Estimators & Filter & Smoother \\ [0.5ex] 
 \hline\hline
 Nexus4 & Nexus4 & 76\% & 83\%  \\ 
 S3 & Nexus4 & 70\% & 77\% \\
 S3mini & Nexus4 & 58\% & 61\% \\
 \hline
 S3 & S3 & 71\% & 80\% \\ 
 Nexus4 & S3 & 71\% & 80\% \\
 S3mini & S3 & 56\% & 64\% \\
 \hline
 S3mini & S3mini & 52\% & 65\% \\ 
 \textbf{Nexus4} & \textbf{S3mini} & \textbf{40\%} & \textbf{58\%} \\
 \textbf{S3} & \textbf{S3mini} & \textbf{44\%} & \textbf{62\%} \\ [1ex]
 \hline
\end{tabular}
\label{table:simHAR}
\end{table}
\section{Conclusion}
The vast majority of collected data is unlabeled. While data-driven algorithms can only learn connections between the observables, estimation of hidden variables of physical or physiological meaning is often required. Model-based estimation algorithms such as the Kalman filter or the particle filter are utilized for real-time estimation and the obtained accuracy is limited due to the use of inaccurate simplified system models. We described a novel methodology and framework that is both model-based and data-driven for utilizing the abundance of available offline data for learning improved causal estimators. We analyzed the use when the misspecified model is a linear model and obtained a bound on the unmodled non-linear behavior energy that enables our learning concept. We provided examples, both for synthetic and real data, where a causal estimator learns from an unlabeled dataset and outperforms the model-based causal estimator.

Future work will include analysis of scenarios when also the system dynamic is misspecified. In such cases only future information that lies within some finite horizon is valid for non-causal estimation, thus, when evaluating the suggested methodology, one should derive the smoothing estimation only by processing a finite size window of measurements.
%
%
%
%
\bibliography{example_paper}
\bibliographystyle{icml2022}
%
%
\newpage
\appendix
\section{Appendix}
\subsection{Kalman smoother block operator}\label{sec:kalblock}
We include here the complete derivation of the Kalman smoother block operator used in Theorem \ref{thm:th3}. Although an elementary result, it does not exist in the literature for smoothing, to the best of our knowledge. We start with formulating the Kalman smoothing problem and follow with the detailed derivation of the block operator.
\subsubsection{Kalman smoother formulation}\label{sec:Appprob}
Assume the discrete-time, linear, finite-dimensional, time-invariant, asymptotically stable system,
\begin{equation}\label{eq:AppSys}
    \begin{split}
        {x}_{k+1} &= A{x}_k + \omega_k\\
        {z}_{k} &= H'{x}_k + v_k,
    \end{split}
\end{equation}
with $x_k \in {\mathbb{R}}^n$, $z_k \in {\mathbb{R}}^m$. The state of the system is $x$, the received measurement is $z$, $\omega$ is a process noise term and $v$ is a measurement noise process. The index $k$ represents discrete time and we are analyzing a finite horizon case such that $0 \leq k < N$. The noise terms $v_k$ and $\omega_k$ are independent, zero mean, Gaussian white processes,
\begin{equation*}
    \operatorname{E}[v_k v_l'] = R \delta_{kl}, \quad
    \operatorname{E}[\omega_k \omega_l'] = Q \delta_{kl}.
\end{equation*}
We assume that $A^{-1}$ exists which is the case if $A$ arises from a real system because then $A$ is the result of a matrix exponential that is always invertible \cite{simon2006optimal}. Suppose that for the estimation of the state ${x}_{k+1}$ (for times $k \geq 0$) based on the measurements ${z}_{0:k}$, the time-invariant Kalman filter is used,
\begin{equation} \label{eq:AppFilter}
    \begin{split}
        \hat{x}_{k+1 \mid k} &= \tilde{A}\hat{x}_{k \mid k-1} + K{z}_k\\
        \hat{x}_{0 \mid -1} &= \hat{x}_0,
    \end{split}
\end{equation}
with $\tilde{A} = A-KH'$, $K=A \bar{\Sigma}H(H' \bar{\Sigma} H + R)^{-1}$ and $\bar{\Sigma}$ is the solution of the steady-state Riccati equation,
\begin{equation} \label{eq:AppBarSigma}
    \bar{\Sigma} = A[\bar{\Sigma} - \bar{\Sigma}H(H' \bar{\Sigma} H + R)^{-1}H'\bar{\Sigma}]A' + Q.
\end{equation}
%
%
We note that for the time-invariant asymptotically stable system \eqref{eq:AppSys} there exists a solution $\bar{\Sigma}$ for equation \eqref{eq:AppBarSigma} that yields $|{\lambda_i(\tilde{F})}| < 1$ (for all $i$), guaranteeing that the filter \eqref{eq:AppFilter} is asymptotically stable. We also note that for time-invariant asymptotically stable systems, a time-varying Kalman filter converges to the the time-invariant Kalman filter in the sense that both the gain $K_k$ and error covariance $\Sigma_k$ asymptotically converge to the values $K$ and $\bar{\Sigma}$.

The estimations of the filter for all $k > 0$ can be written as,
\begin{equation}\label{eq:filterGeneral}
\begin{split}
        \hat{x}_{k \mid k-1} &= \tilde{A}^k\hat{x}_{0 \mid -1} + \sum_{i=0}^{k-1} {\tilde{A}}^i K z_{k-i-1}\\
    &= \tilde{A}^k\hat{x}_{0 \mid -1} + \sum_{i=0}^{k-1} {\tilde{A}}^{k-i-1} K z_{i},
\end{split}
\end{equation}
which can trivially be written in a block-matrix form,
\begin{equation}\label{eq:filterBlock}
    \xi_N = \bar{\Xi}_N Z_N,
\end{equation}
where $\xi_N, Z_N \in {\mathbb{R}^{nN}}$ are the input-output block vectors,
\begin{equation}
    \begin{split}
        Z_N &= \begin{bmatrix}
        z_0' & z_1' & \dots & z_{N-1}',
    \end{bmatrix} '\\
        \xi_N &= \begin{bmatrix}
        \hat{x}_{0 \mid -1}' & \hat{x}_{1 \mid 0}' & \dots & \hat{x}_{N-1 \mid N-2}'
    \end{bmatrix} '        
    \end{split}
\end{equation}
and $\bar{\Xi}_N$ is the operator block matrix of the filter with dimensions $[nN \times nN]$ and having the entries
\begin{equation}
    \bar{\Xi}_{N}[r,c] = {\mathcal I}(c < r) \tilde{A}^{r-1-c}K.
\end{equation}

Suppose that for a non-causal estimation of the state $x_j$ the time-invariant Kalman smoother is used such that for times $k \geq j$ the estimation of $x_j$ using measurements up to and including time $k$, $z_{0:k}$, is given by the recursive relation,
\begin{equation}\label{eq:smoother}
    \hat{x}_{j \mid k} = \hat{x}_{j \mid k-1} + K^a_{k-j}\tilde{z}_k.
\end{equation}
with the gain matrix given by,
\begin{equation}\label{eq:smootherGain}
    K^a_{k-j} = \Bar{\Sigma} \tilde{A}^{k-j'} [A \bar{\Sigma}]^{-1}K.
\end{equation}
The fixed point smoother is driven from the innovations process $\tilde{z}_k = (z_k - H'\hat{x}_{k \mid k-1})$ of the Kalman filter. We refer the reader to Section 7.2 in \cite{anderson2012optimal} for the complete derivation of fixed-point smoothing. We analyze the time-invariant estimators but would like to point out that under the assumption
\begin{equation}\label{eq:simplicity}
    \Sigma_{0} = \bar{\Sigma},
\end{equation}
they coincide with the time variant optimal estimators. With respect to the convergence property of the time variant Kalman filter, we note that our setup is also suitable for a time varying smoother such that the corresponding filter was utilized at time $k=k_0 \ll 0$ such that at time $k=0$ the gain $K_k$ and error covariance $\Sigma_k$ have converged, $\|K_k-K\|_2 \approx 0$, $\|\Sigma_k-\bar{\Sigma}\|_2 \approx 0$. \
For a detailed analysis of time-invariant Kalman filters see Chapter 4 in \cite{anderson2012optimal}.

Let $\xi^s_N \in {\mathbb{R}^{nN}}$ be the block vector,
\begin{equation}
    \xi^s_N = \begin{bmatrix}
        \hat{x}_{0 \mid N-1}' & \hat{x}_{1 \mid N-1}' & \dots & \hat{x}_{N-1 \mid N-1}'.
    \end{bmatrix} '        
\end{equation}
We are looking for the time-invariant smoother block operator, $\bar{\Xi}^s_N$, such that,
\begin{equation}\label{eq:smootherOut}
    \xi^s_N = \bar{\Xi}^s_N Z_N.
\end{equation}
For simplicity we assume $\hat{x}_{0 \mid -1}=0$ but emphasize that the derivation of $\bar{\Xi}^s_N$ is performed in the exact same manner when including the term for $\hat{x}_{0 \mid -1}$.
\subsubsection{Kalman smoother block operator}
The smoother block operator is a matrix of dimensions $[nN \times nN]$ having the entries,
\begin{equation}
    \bar{\Xi}^s_{N}[r,c] = {\tilde{G}_{r,c}}K,
\end{equation}
where 
\begin{equation}
    \begin{split}
        &\tilde{B}_{k,i} =  \tilde{A}^{k-i-1}
        - \Bar{\Sigma}  \tilde{D}_{k,i,k} \\
        &\tilde{C}_{k,i} = \Bar{\Sigma} (\tilde{A}^{i-k'} [A \Bar{\Sigma}]^{-1} 
        - \tilde{D}_{k,i,i+1})\\
        &\tilde{D}_{k,i,m} = \sum_{n=m}^{N-1} \tilde{E}_{k,i,n}\\
        &\tilde{E}_{k,i,n} = \tilde{A}^{n-k'} [A \Bar{\Sigma}]^{-1} K H'\tilde{A}^{n-i-1}\\
        &\tilde{G}_{k,i} = {\cal I}(i<k){\tilde{B}_{k,i}} + {\cal I}(i \geq k){\tilde{C}_{k,i}}.
    \end{split}
\end{equation}
\subsubsection{Proof}
From the recursive equation of the smoother \eqref{eq:smoother}, the estimations of the smoother for all $k \geq 0$ can be written as,
\begin{equation}\label{eq:smootherGeneral}
    \hat{x}_{k \mid N-1} = \hat{x}_{k \mid k-1} + \sum_{i=k}^{N-1}K^a_{i-k}(z_i - H'\hat{x}_{i \mid i-1}). 
\end{equation}
This expression is a function both of the measurements and the filter estimations. First we derive an expression that is a direct function of the measurements only,
\begin{equation}\label{eq:startSmoothingFromAllMeas}
    \begin{split}
        &\hat{x}_{k \mid N-1} = \hat{x}_{k \mid k-1} + \sum_{i=k}^{N-1} K^a_{i-k} (z_i - H'\hat{x}_{i \mid i-1})\\ 
        &\overset{(i)}{=} \sum_{i=0}^{k-1}\tilde{A}^{k-1-i}Kz_i + \sum_{i=k}^{N-1} K^a_{i-k} (z_i - H'\hat{x}_{i \mid i-1}) \\
        &= \sum_{i=0}^{k-1}\tilde{A}^{k-1-i}Kz_i +  \sum_{i=k}^{N-1} K^a_{i-k} z_i - \sum_{i=k}^{N-1} K^a_{i-k} H'\hat{x}_{i \mid i-1}\\
        &\overset{(ii)}{=} \sum_{i=0}^{k-1}\tilde{A}^{k-1-i}Kz_i +  \sum_{i=k}^{N-1} K^a_{i-k} z_i\\
        &- \sum_{i=k}^{N-1} K^a_{i-k} H'\sum_{m=0}^{i-1}\tilde{A}^{i-1-m}Kz_{m}\\
        &= \sum_{i=0}^{k-1}\tilde{A}^{k-1-i}Kz_i +  \sum_{i=k}^{N-1} K^a_{i-k} z_i\\
        &- \sum_{n=k}^{N-1} K^a_{n-k} H'\sum_{i=0}^{n-1}\tilde{A}^{n-i-1}Kz_i\\
    \end{split}
\end{equation}
where in $(i)$ we substituted equation \eqref{eq:filterGeneral} for $\hat{x}_{k \mid k-1}$ and in $(ii)$ we substituted it for $\hat{x}_{i \mid i-1}$. Next we manipulate the last term in equation \eqref{eq:startSmoothingFromAllMeas} into a sum of two terms, a causal term that includes measurements up to and including time $k-1$ and a non-causal term that includes measurements from time $k$ and up to and including time $N-1$,
\begin{equation}\label{eq:dev1}
    \begin{split}
        &\sum_{n=k}^{N-1} K^a_{n-k} H'\sum_{i=0}^{n-1}\tilde{A}^{n-1-i}Kz_i \\
        &= \sum_{n=k}^{N-1} K^a_{n-k} H'\left(\sum_{i=0}^{k-1}\tilde{A}^{n-1-i}Kz_i + \sum_{i=k}^{n-1}\tilde{A}^{n-1-i}Kz_i\right)\\
        &= \sum_{n=k}^{N-1} \sum_{i=0}^{k-1}K^a_{n-k} H'\tilde{A}^{n-1-i}Kz_i\\
        &+ \sum_{n=k}^{N-1} \sum_{i=k}^{n-1}K^a_{n-k} H'\tilde{A}^{n-1-i}Kz_i\\
        &\overset{(i)}{=} \sum_{i=0}^{k-1} \sum_{n=k}^{N-1} K^a_{n-k} H'\tilde{A}^{n-1-i}Kz_i\\
        &+ \sum_{n=k+1}^{N-1} \sum_{i=k}^{n-1}K^a_{n-k} H'\tilde{A}^{n-1-i}Kz_i
    \end{split}
\end{equation}
where in $(i)$ we switched the summation order and noted that the summation in the second term actually starts at n=$k+1$. Noting that,
\begin{equation}
    \begin{split}
        &\sum_{n=k+1}^{N-1} \sum_{i=k}^{n-1} \phi_{n,i}\\
        &=\sum_{n=k+1}^{N-1} \phi_{n,k} + \phi_{n,k+1} + \dots + \phi_{n,n-1}\\\\
        &\overset{(i)}{=} \phi_{k+1,k}\\
        &+ \phi_{k+2,k} + \phi_{k+2,k+1}\\ 
        &+ \phi_{k+3,k} + \phi_{k+3,k+1} + \phi_{k+3,k+2}\\
        &\vdots\\
        &+ \phi_{N-1,k} + \phi_{N-1,k+1} + \phi_{N-1,k+2} + \dots + \phi_{N-1,N-2}\\\\
        &\overset{(ii)}{=} \sum_{i=k}^{N-2}\sum_{n=i+1}^{N-1}\phi_{n,i}
    \end{split}
\end{equation}
where in $(ii)$ the internal sum is over the columns in $(i)$ and the outer sum over rows, substituting $\phi_{n,i} = K^a_{n-k} H'\tilde{A}^{n-1-i}Kz_i$ we can write equation \eqref{eq:dev1} as,
\begin{equation}\label{eq:dev2}
    \begin{split}
        &\sum_{n=k}^{N-1} K^a_{n-k} H'\sum_{i=0}^{n-1}\tilde{A}^{n-1-i}Kz_i \\
        &= \sum_{i=0}^{k-1} \sum_{n=k}^{N-1} K^a_{n-k} H'\tilde{A}^{n-1-i}Kz_i\\
        &+ \sum_{i=k}^{N-2}\sum_{n=i+1}^{N-1}K^a_{n-k} H'\tilde{A}^{n-1-i}Kz_i.
    \end{split}
\end{equation}
Now by substituting this expression into equation \eqref{eq:startSmoothingFromAllMeas} we obtain $\hat{x}_{k \mid N-1}$ as a sum of a {\color{black}{causal}} and a {\color{black}{non-causal}} term,
\begin{equation}\label{eq:dev3}
    \begin{split}
        &\hat{x}_{k \mid N-1} = {\color{black}{\hat{x}^-_{k \mid N-1}}} + {\color{black}{\hat{x}^+_{k \mid N-1}}},
    \end{split}
\end{equation}        
where
\begin{equation}\label{eq:dev300}
    \begin{split}
        &{\color{black}{\hat{x}^-_{k \mid N-1}}} 
        = {\color{black}{\sum_{i=0}^{k-1}\tilde{A}^{k-1-i}Kz_i
        -  \sum_{i=0}^{k-1} \sum_{n=k}^{N-1} K^a_{n-k} H'\tilde{A}^{n-1-i}Kz_i}}\\
        &{\color{black}{\hat{x}^+_{k \mid N-1}}} ={\color{black}{\sum_{i=k}^{N-1} K^a_{i-k} z_i - \sum_{i=k}^{N-2}\sum_{n=i+1}^{N-1}K^a_{n-k} H'\tilde{A}^{n-1-i}Kz_i}}.
    \end{split}
\end{equation}        
We separately further develop each term. Starting with the causal term, simple algebraic manipulations yield,
\begin{equation}\label{eq:dev31_causal}
    \begin{split}
        &{\color{black}{\hat{x}^-_{k \mid N-1}}}  =\\
        &{\color{black}{=\sum_{i=0}^{k-1}\tilde{A}^{k-1-i}Kz_i
        -  \sum_{n=k}^{N-1} \sum_{i=0}^{k-1} K^a_{n-k} H'\tilde{A}^{n}\tilde{A}^{-1-i}Kz_i}}\\
        &={\color{black}{\tilde{A}^{k}\sum_{i=0}^{k-1}\tilde{A}^{-1-i}Kz_i
        -  \sum_{n=k}^{N-1} K^a_{n-k} H'\tilde{A}^{n} \sum_{i=0}^{k-1} \tilde{A}^{-1-i}Kz_i}}\\
        &= {\color{black}{\left(\tilde{A}^{k}
        -  \sum_{n=k}^{N-1} K^a_{n-k} H'\tilde{A}^{n} \right)\sum_{i=0}^{k-1} \tilde{A}^{-1-i}Kz_i}}.
    \end{split}
\end{equation}        
Next, Substituting the gain from equation \eqref{eq:smootherGain}, $K^a_{n-k} = \Bar{\Sigma} \tilde{A}^{n-k'} [A \bar{\Sigma}]^{-1}K$ the causal term becomes,  
\begin{equation}\label{eq:dev4_causal}
    \begin{split}        
        &\hat{x}^-_{k \mid N-1} =\\
        &= {\color{black}{\left(\tilde{A}^{k}
        -  \sum_{n=k}^{N-1} \Bar{\Sigma} \tilde{A}^{n-k'} [A \bar{\Sigma}]^{-1}K H'\tilde{A}^{n} \right)\sum_{i=0}^{k-1} \tilde{A}^{-1-i}Kz_i}}\\ 
        &= {\color{black}{\left(\tilde{A}^{k-1}
        -  \Bar{\Sigma} \sum_{n=k}^{N-1} \tilde{A}^{n-k'} [A \bar{\Sigma}]^{-1}K H'\tilde{A}^{n-1} \right)\sum_{i=0}^{k-1} \tilde{A}^{-i}Kz_i}}.
    \end{split}
\end{equation}
Define the matrices,
\begin{equation}
    \begin{split}
        &\tilde{B}_{k,i} =  \tilde{A}^{k-i-1}
        - \Bar{\Sigma}  \tilde{D}_{k,i,k} \\
        &\tilde{C}_{k,i} = \Bar{\Sigma} (\tilde{A}^{i-k'} [A \Bar{\Sigma}]^{-1} 
        - \tilde{D}_{k,i,i+1})\\
        &\tilde{D}_{k,i,m} = \sum_{n=m}^{N-1} \tilde{E}_{k,i,n}\\
        &\tilde{E}_{k,i,n} = \tilde{A}^{n-k'} [A \Bar{\Sigma}]^{-1} K H'\tilde{A}^{n-i-1}\\
        &\tilde{G}_{k,i} = {\cal I}(i<k){\tilde{B}_{k,i}} + {\cal I}(i \geq k){\tilde{C}_{k,i}},
    \end{split}
\end{equation}
and substitute $\tilde{E}_{k,i,n}$ into \eqref{eq:dev4_causal},
\begin{equation}\label{eq:dev5}
    \begin{split} 
        &\hat{x}^-_{k \mid N-1} =\\
        &= {\color{black}{\left(\tilde{A}^{k-1}
        -  \Bar{\Sigma} \sum_{n=k}^{N-1} \tilde{E}_{k,0,n} \right)\sum_{i=0}^{k-1} \tilde{A}^{-i}Kz_i}}\\
        &= {\color{black}{\sum_{i=0}^{k-1} \left(\tilde{A}^{k-i-1}
        -  \Bar{\Sigma} \sum_{n=k}^{N-1} \tilde{E}_{k,0,n} \tilde{A}^{-i} \right) Kz_i}}\\
        &= {\color{black}{\sum_{i=0}^{k-1} \left(\tilde{A}^{k-i-1}
        -  \Bar{\Sigma} \sum_{n=k}^{N-1} \tilde{E}_{k,i,n}\right) Kz_i}}.
    \end{split}
\end{equation}
Now, substituting $\tilde{D}_{k,i,m}$ and $\tilde{B}_{k,i,m}$ we obtain,
\begin{equation}\label{eq:dev6}
    \begin{split} 
    &\hat{x}^-_{k \mid N-1}
        = {\color{black}{\sum_{i=0}^{k-1} \left(\tilde{A}^{k-i-1}
        -  \Bar{\Sigma} \tilde{D}_{k,i,k}\right) Kz_i}} = {\color{black}{\sum_{i=0}^{k-1} \tilde{B}_{k,i} Kz_i}}.
    \end{split}
\end{equation}
In a similar manner we develop the non-causal term, starting with simple algebraic manipulations,
\begin{equation}\label{eq:dev31_non}
    \begin{split}
        &\hat{x}^+_{k \mid N-1}
        = {\color{black}{\sum_{i=k}^{N-1} K^a_{i-k} z_i - \sum_{i=k}^{N-2}\sum_{n=i+1}^{N-1}K^a_{n-k} H'\tilde{A}^{n-1-i}Kz_i}}\\
        &={\color{black}{  \sum_{i=k}^{N-2} K^a_{i-k} z_i
        - \sum_{i=k}^{N-2}\sum_{n=i+1}^{N-1}K^a_{n-k}H'\tilde{A}^{n-1-i}Kz_i}} {\color{black}{+  z^{(K)}_{N-1}}}\\
        &={\color{black}{  \sum_{i=k}^{N-2}\left( K^a_{i-k}  - \sum_{n=i+1}^{N-1}K^a_{n-k}H'\tilde{A}^{n-1-i}K\right)z_i + z^{(K)}_{N-1}}},
    \end{split}
\end{equation}        
Where we defined $z^{(K)}_{N-1} = K^a_{N-1-k} z_{N-1}$. Continuing by substitute the gain,   
\begin{equation}\label{eq:dev4_non}
    \begin{split}        
        &\hat{x}^+_{k \mid N-1} 
        =\sum_{i=k}^{N-2}\Bar{\Sigma} \tilde{A}^{i-k'} [A \bar{\Sigma}]^{-1}Kz_i\\
        &- \sum_{i=k}^{N-2}\sum_{n=i+1}^{N-1}\Bar{\Sigma} \tilde{A}^{n-k'} [A \bar{\Sigma}]^{-1}KH'\tilde{A}^{n-1-i}Kz_i\\\\
        &+  \Bar{\Sigma} \tilde{A}^{N-1-k'} [A \bar{\Sigma}]^{-1}K z_{N-1}\\\\
        &=\sum_{i=k}^{N-1} \Bar{\Sigma} \tilde{A}^{i-k'} [A \bar{\Sigma}]^{-1}Kz_i\\
        &- \sum_{i=k}^{N-1} \Bar{\Sigma} \sum_{n=i+1}^{N-1} \tilde{A}^{n-k'} [A \bar{\Sigma}]^{-1}KH'\tilde{A}^{n-1-i}Kz_i,
    \end{split}
\end{equation}
then substituting $\tilde{E}_{k,i,n}$ into \eqref{eq:dev4_non},
\begin{equation}\label{eq:dev5_non}
    \begin{split} 
        &\hat{x}^+_{k \mid N-1} 
        =\sum_{i=k}^{N-1} \Bar{\Sigma} \left(  \tilde{A}^{i-k'} [A \bar{\Sigma}]^{-1}  -  \sum_{n=i+1}^{N-1} \tilde{E}_{k,i,n}\right)Kz_i,
    \end{split}
\end{equation}
and finally substituting $\tilde{D}_{k,i,m}$ and $\tilde{C}_{k,i}$,
\begin{equation}\label{eq:dev6}
    \begin{split} 
    \hat{x}^+_{k \mid N-1} 
        &=\sum_{i=k}^{N-1}\Bar{\Sigma} \left(  \tilde{A}^{i-k'} [A \bar{\Sigma}]^{-1}  -  \tilde{D}_{k,i,i+1}\right)Kz_i\\
        &= \sum_{i=k}^{N-1}\tilde{C}_{k,i}Kz_i.
    \end{split}
\end{equation}
Now we combine the causal and non-causal terms yielding,
\begin{equation}\label{eq:dev7}
    \begin{split} 
    &\hat{x}_{k \mid N-1} = \hat{x}^-_{k \mid N-1} + \hat{x}^+_{k \mid N-1}\\
     &={\color{black}{\sum_{i=0}^{k-1} \tilde{B}_{k,i} Kz_i}}
        {\color{black}{+  \sum_{i=k}^{N-1} \tilde{C}_{k,i}Kz_i}} = \sum_{i=0}^{N-1}\tilde{G}_{k,i} Kz_i.
    \end{split}
\end{equation}
From equation \eqref{eq:dev7} it is immediate to deduce the $[nN \times nN]$ smoothing block operator having the entries
\begin{equation}
    \bar{\Xi}^s_{N}[r,c] = {\tilde{G}_{r,c}}K,
\end{equation}
such that
\begin{equation*}
    \xi^s_N = \bar{\Xi}^s_N Z_N.
\end{equation*}
%
%
\subsection{Human activity recognition - technical details}\label{sec:HarTech}
We detail here the architecture of the filter we utilized for real time estimation of the human action given past observations. The smoother has a similar architecture detailed below.

The raw observations ($z_{k},\tilde{z}_k \in {\mathbb{R}}^3$) are three dimensional accelerometer measurements which are recieved from different smartphones at different sampling rates (see \cite{stisen2015smart} for details). We interpolate them to a common sampling rate of $200$ Herz as as to utilize the exact same estimator architecture for all smartphones. To preserve causality, the interpolation used is zero-order-hold \cite{oppenheim1975digital}. Given raw data $r_n$ at sampling rate $r_s$ Herz, resampling it to a sampling rate $f_s$ Herz using zero-order-hold interpolation is expressed by
\begin{equation*}
    \rho_k = r_n, \quad n\frac{r_s}{f_s} \leq k \leq (n+1)\frac{r_s}{f_s}.
\end{equation*}

We utilize the DeepConvLSTM architecture presented in \cite{ordonez2016deep} with slight changes. Four sequential convolution layers are processing finite equal length windows containing four seconds of the observed raw measurements (after the resampling to $200$ Herz), and produce for each processed window a 12-dimensional vector of extracted learned features. The windows are overlapping by $3.75$ seconds and thus the 12-dimensional feature vector is at a rate of $4$ Herz. A single window containing four seconds of raw measurements is given by $z_{k-d f_s+1:k} \in {\mathbb{R}}^{m d f_s \times 1}$ with $m=3$, $d=4$ [sec] and $f_s=200$ [Herz]. The output of the first convolution layer is given by,
\begin{equation*}
    c^{(0)}_k = \sigma\left(W^{(0)} \cdot z_{lk-d f_s+1:lk} + b^{(0)}\right),
\end{equation*}
where $W^{(0)} \in {\mathbb{R}}^{j \times m d f_s}$ with $j=12$ is a learned matrix, $b^{(0)} \in {\mathbb{R}}^{j \times 1}$ is a learned bias vector, $l=50$ reflects the overlap of $3.75$ seconds between sequential windows and $\sigma(\cdot)$ is ReLU activation. We set $k=0$ as the starting index of each input time-series and set $z_k =0$ for all $k < 0$. For $i=1,2,3$, the output of the following convolution layers is given by,
\begin{equation*}
    c^{(i)}_k = \sigma\left(W^{(i)} \cdot c^{(i-1)}_{k-d\tilde{f}_s+1:k} + b^{(i)}\right),
\end{equation*}
where $\tilde{f}_s=f_s/l=4$ [Herz], $c^{(i-1)}_{k-d\tilde{f}_s+1:k} \in {\mathbb{R}}^{jd\tilde{f}_s \times 1}$ is the output of convolution layer $i-1$, $W^{(i)} \in {\mathbb{R}}^{j \times j d \tilde{f}_s}$ are learned matrices and $b^{(i)} \in {\mathbb{R}}^{j \times 1}$ are learned bias vectors. 

A single layer LSTM with a hidden size $h=30$ processes the 12-dimensional time-series $c^{(3)}_k$ and its output is a time-series $o^{(f)}_k \in {\mathbb{R}}^h$ which is connected to a linear layer that maps the 30-dimensional LSTM state to a vector whose dimension equals the number of categories (6). Finally a log-softmax layer outputs the time-series $\hat{x}_{k+1 \mid k} \in {\mathbb{R}}^6$ (at $4$ Herz) of log-probabilities of each category. See \citet{zhang2020dive} for a detailed explanation of convolution layers LSTMs, softmax layers and ReLU activation.

The smoother utilizes a single bi-directional LSTM and so outputs the non-causal estimation. The forward and backward layers both have the same hidden size $h=30$ where the backward layer processes the measurements in a reversed order. The output of the bi-directional LSTM is a time-series $[o^{(f)}_k, o^{(b)}_k]  \in {\mathbb{R}}^{2h}$ which is connected to a linear layer that maps the 60-dimensional bi-directional LSTM state to a vector whose dimension equals the number of categories (6). Finally a log-softmax layer outputs the time-series $\hat{x}_{k \mid N-1} \in {\mathbb{R}}^6$ (at $4$ Herz) of log-probabilities of each category, where $N$ is the length of the time-series.

We note that the architecture presented by \cite{ordonez2016deep} consists of two LSTM layers while we use a single layer as suggested recently by \cite{bock2021improving} for the HAR task. We train the estimators using the Adam optimizer \cite{kingma2014adam} with a learning rate of $0.001$ and average the results over 10 independent train/test splits of the dataset. 
\end{document}